\documentclass[letterpaper, 10 pt, conference]{ieeeconf}  

\IEEEoverridecommandlockouts                              

\usepackage{graphicx}
\usepackage{amsmath}
\usepackage{amssymb}
\usepackage{amsfonts}
\usepackage{booktabs}
\usepackage{multirow}
\usepackage{tabularx}
\let\BORAlabelindent\labelindent
\let\labelindent\relax
\usepackage{enumitem}
\let\labelindent\BORAlabelindent

\usepackage{xcolor}
\usepackage{placeins}
\title{\LARGE \bf
BORA: Bridging Offline Reinforcement Learning and Online Residual Adaptation for Real-World Dexterous VLA Models
}

\author{
Zhongxi Chen$^{1*}$ \quad
Yifan Han$^{2*}$ \quad
Bin Qiu$^{1*}$ \quad
Zhangliang Gao$^{1}$\\
Yanming Shao$^{3}$ \quad
Huanming Liu$^{4}$ \quad
Congsheng Xu$^{1}$ \quad
Xiaoyu Chen$^{1}$\\
Xingyu Ye$^{5}$ \quad
Yao Mu$^{1\dagger}$ \quad
Wenzhao Lian$^{1\dagger}$\\[1mm]
$^{1}$Shanghai Jiao Tong University \quad
$^{2}$CASIA \quad
$^{3}$Shanghai AI Laboratory\\
$^{4}$USTC \quad
$^{5}$Duke University\\[1mm]
{\small $*$ Co-first authors \qquad $\dagger$ Corresponding authors}
}

\makeatletter
\IEEEaftertitletext{%
\begin{minipage}{\textwidth}
\centering
\includegraphics[width=0.94\textwidth]{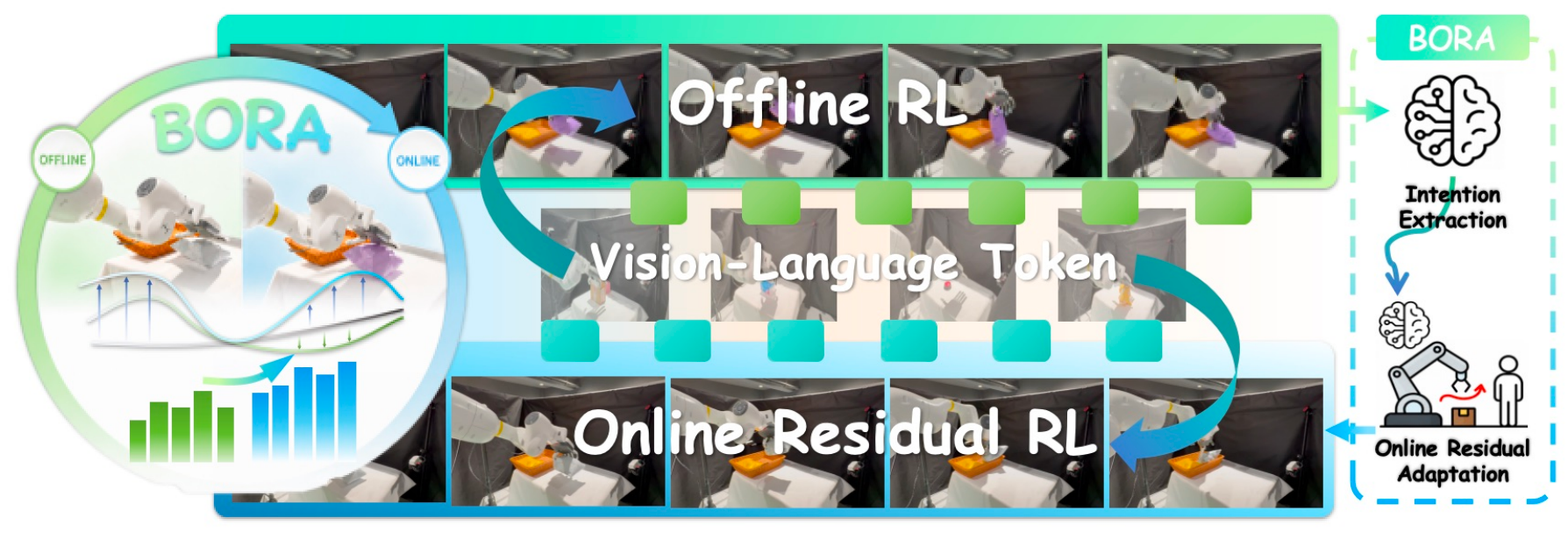}
\def\@captype{figure}
\caption{BORA connects offline policy learning, human intervention, and online residual adaptation. Human corrections collected during execution update a residual actor while the VLA base remains fixed.}
\label{fig:teaser}
\end{minipage}\vspace{1em}}
\makeatother

\begin{document}

\maketitle
\thispagestyle{empty}
\pagestyle{empty}

\begin{abstract}
Vision-Language-Action (VLA) policies provide strong behavioral priors for dexterous manipulation, yet adapting them on real robots remains challenging because high-DoF contact failures are difficult for humans to correct and online interaction is expensive. We present BORA, an offline-to-online reinforcement learning system that integrates an action-conditioned critic into a consistency-policy VLA and reuses the learned critic for frozen-base residual adaptation. To obtain executable corrective data, BORA combines wearable arm--hand teleoperation with a demonstration-guided local policy that translates coarse human intent into coordinated, embodiment-specific finger motions for contact-rich skills. Online robot rollouts and human corrections are mixed with offline data to update only a lightweight residual actor, avoiding full-model fine-tuning. We evaluate BORA on six real-world tasks using single-arm and bimanual platforms equipped with two dexterous-hand models. With only 20 online trajectories per task, BORA improves average success from 60.8\% to 82.5\% on standard objects and from 52\% to 70\% on held-out objects, while policy assistance substantially improves intervention reliability in bimanual twisting. These results demonstrate a practical route from executable human correction to efficient real-robot VLA adaptation.
\end{abstract}


\section{Introduction}
\label{sec:intro}

Vision-Language-Action (VLA) models offer a promising route to generalizable robot control, yet real-world dexterous manipulation remains difficult because its high-dimensional action space couples arm, wrist, and multi-finger motion. A task may admit many valid grasps and finger trajectories, but small deviations can change the contact mode and turn success into failure. Offline demonstrations provide useful behavioral priors while covering only a small fraction of feasible contact configurations and recovery behaviors. Deployment therefore demands physical adaptation beyond offline imitation.

Offline-to-online reinforcement learning offers a natural mechanism for such adaptation, but most real-robot systems focus on comparatively low-dimensional arm or parallel-gripper control~\cite{chen2025conrft,xu2026rl,chen2025pirl,nakamoto2023cal,ball2023efficient,lei2025rl}. Extending this paradigm to an arm--dexterous-hand system exposes two coupled bottlenecks. On the learning side, policy and value estimation must operate over high-dimensional, multimodal action chunks under sparse task feedback. On the interaction side, the system needs infrastructure that can produce reliable, coordinated arm--hand corrections when the policy enters contact states outside the offline distribution.

The intervention bottleneck is most acute in contact-intensive tasks. Direct teleoperation requires the operator to detect an impending failure while coordinating the arm, wrist, and multiple fingers, imposing substantial perceptual and motor load. Twisting makes the embodiment gap explicit: differences in thumb placement, workspace, and motion prevent human fingertip poses from reliably reproducing the opposing contacts and tangential forces needed for rotation. A failed takeover both aborts the rollout and yields a poor corrective trajectory. Dexterous offline-to-online RL therefore needs task-aware assistance that converts human intent into executable robot motion instead of relying on pose retargeting alone.

Learning from these corrections presents a complementary challenge. Offline demonstrations may be limited or suboptimal~\cite{mandlekar2021matters}, while friction, contact dynamics, and hand--object occlusion induce deployment states absent from the dataset. Repeated full-parameter VLA fine-tuning is expensive and may disrupt pretrained representations as the online distribution changes~\cite{zhou2025efficient}. Conversely, an externally attached critic may evaluate visual and state features that are not aligned with the VLM cognition tokens and action chunks driving the policy. A practical system should preserve the offline policy as a stable prior, ground value estimation in its VLM representation, and learn compact corrections from heterogeneous online experience.

We address these requirements with \textbf{BORA (Bridging Offline RL and Online Residual Adaptation)}, an integrated offline-to-online RL system for dexterous VLA policies (Fig.~\ref{fig:teaser}). Offline, an action-conditioned critic evaluates action chunks using the same VLM cognition tokens as a consistency-policy actor. Online, BORA freezes the base VLA, inherits the critic, and updates a lightweight residual actor from mixed offline and online replay. Two intervention modes complete the loop: wearable teleoperation corrects general arm--hand failures, while local policy assistance maps human rotation intent to embodiment-specific finger coordination for twisting. BORA thus connects offline value learning, executable physical correction, and efficient online adaptation within one system.

Our main contributions are threefold:
\begin{itemize}[nosep,leftmargin=*]
\item \textbf{Integrated offline-to-online RL for dexterous VLAs:} An action-conditioned critic shares the actor's VLM representation during offline policy improvement and is inherited online to guide a residual actor while the base VLA remains frozen.
\item \textbf{Executable high-DoF intervention:} BORA combines the wearable arm--hand interface adopted from DexHiL~\cite{han2026dexhil} with local policy assistance that translates human intent into coordinated, embodiment-specific contact corrections.
\item \textbf{Cross-hardware real-robot evaluation:} Six tasks on Franka--DexHand021 and bimanual ROKAE--Linker Hand L20 platforms evaluate offline learning, online adaptation, held-out-object generalization, and assisted intervention.
\end{itemize}

\section{Related Work}

\paragraph{Vision-Language-Action Models for Dexterous Manipulation}
VLA models adapt representations from pretrained vision-language models to robot control through action heads. Earlier models use tokenized or transformer-based action generation~\cite{zitkovich2023rt,mees2024octo,kim2024openvla}, while continuous policies use diffusion or flow-based action heads~\cite{li2025scalable,chen2025flowing,black2024pi_0,intelligence2025pi_}. Consistency policies support few-step action generation~\cite{prasad2024consistency,lu2024manicm}. Dexterous VLA and vision-language-grasp models extend these capabilities to multi-fingered manipulation~\cite{luo2026being,zhong2025dexgraspvla,he2025dexvlg}. Being-H0~\cite{luo2025being} and VITRA~\cite{li2025scalable} study pretraining from human data, while Being-H0.5~\cite{luo2026being} explores cross-embodiment generalization. BORA builds on a pretrained VLA and studies its subsequent adaptation to real-world dexterous execution.

\paragraph{Human Intervention and Policy-Assisted Teleoperation}
Human intervention and teleoperation provide an important interface for dexterous manipulation~\cite{mandlekar2020human}. GeoRT~\cite{yin2025geometric} retargets human finger motion to robot-hand configurations, while DexHiL~\cite{han2026dexhil} enables coordinated arm--hand intervention for online correction. Direct retargeting nevertheless requires the operator to generate fine-grained contact behavior, making regrasping, finger gaiting, and object reorientation difficult. Recent methods therefore use learned local policies to execute low-level skills from coarse human motion or intent; DexGen~\cite{yin2025dexteritygen} and TeleDexter~\cite{li2026towards} follow this direction. CHORD~\cite{zhu2026learning} further shows that object-centric contact wrenches capture how contact changes object motion more directly than contact positions alone. Motivated by this observation, we derive progress-aligned force intent from fingertip motion and object dynamics to guide an embodiment-specific local policy.

\paragraph{Post-Training and Adaptation of VLA Policies}
Offline post-training reuses collected robot data but remains limited by demonstration quality and train--deployment mismatch~\cite{huang2025co}. Online RL instead improves performance through interaction with the target environment~\cite{chen2025conrft,xu2026rl,chen2025pirl}, while offline-to-online methods reuse prior data or value estimates to initialize adaptation~\cite{nakamoto2023cal,ball2023efficient}. Residual RL constrains this adaptation to corrections around an existing policy~\cite{ankile2025imitation,su2026rfs}. In dexterous manipulation, DexHiL~\cite{han2026dexhil} combines coordinated arm--hand intervention with intervention-aware sampling and supervised policy updates. BORA retains its wearable intervention interface for the five single-arm tasks, adds local assistance for bimanual twisting, and couples both modes to offline-initialized value guidance and online residual learning.

\section{Method}
\label{sec:method}

\begin{figure*}[t]
\centering
\includegraphics[width=0.98\textwidth]{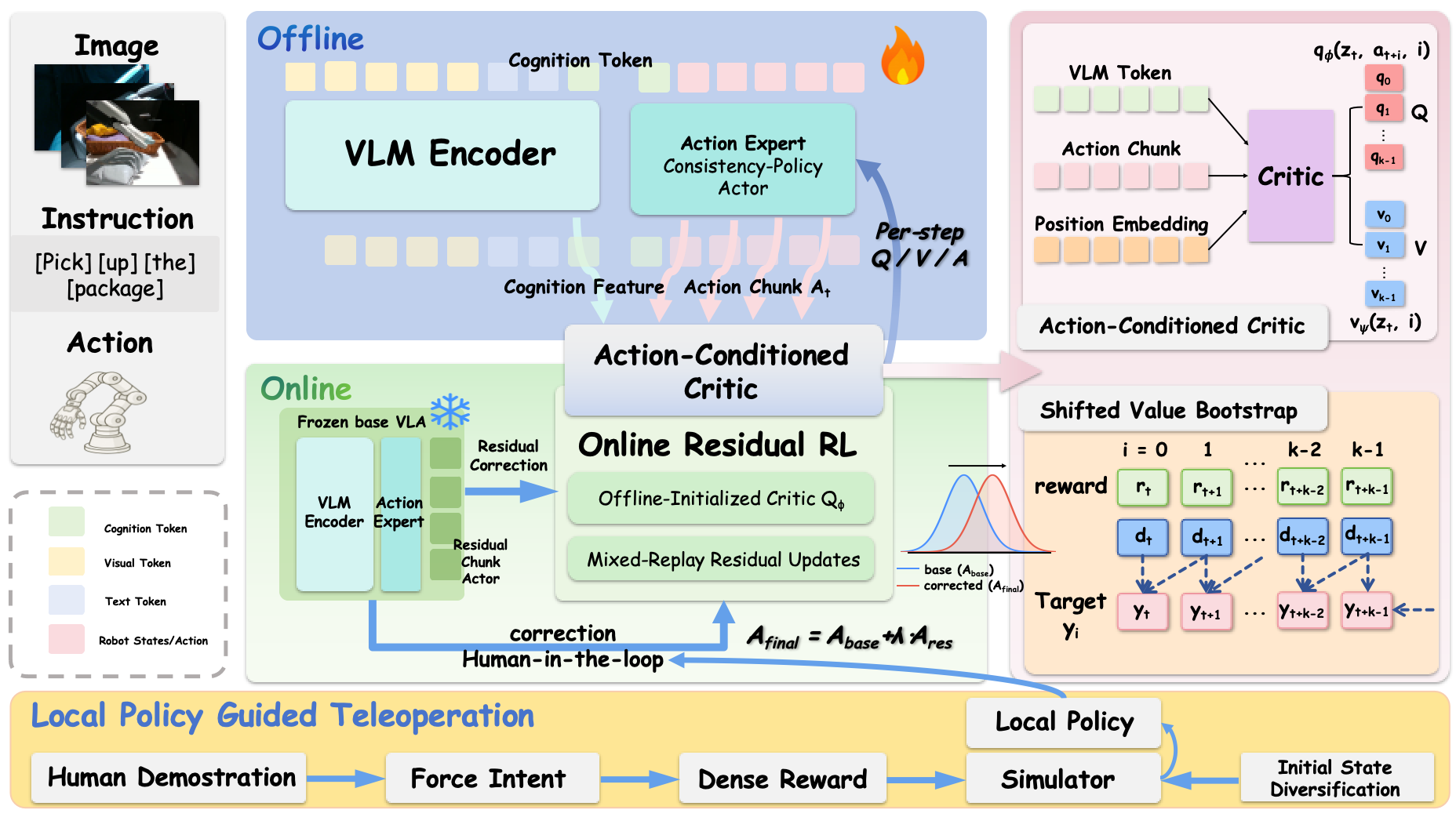}
\caption{\textbf{Overview of BORA.} Offline, multimodal observations and actions train a VLM-based consistency actor and an action-conditioned critic; shifted bootstrap targets provide per-step value guidance. Online, the offline-initialized critic guides residual chunk updates from robot interaction and human corrections while the base VLA remains frozen, yielding $A_{\mathrm{final}}=A_{\mathrm{base}}+\lambda A_{\mathrm{res}}$. For contact-intensive twisting, human demonstrations define force intent and dense simulation rewards for a local policy that converts operator intent into executable corrections.}
\label{fig:double}
\end{figure*}

\paragraph{System Overview}
BORA comprises offline policy and value initialization, human intervention, and online learning from robot interaction and corrections (Fig.~\ref{fig:double}). The offline policy supplies action chunks; when correction is needed, the operator intervenes through direct teleoperation or local assistance. The resulting transitions enter replay to update a residual actor, whose corrections modify subsequent actions. The intervention subsystem makes corrections executable, while the learning pipeline converts them into improved behavior.

\subsection{Offline Policy and Value Initialization}
\label{subsec:offline_rl}

The VLA actor uses a consistency policy to generate continuous action chunks in 1--3 denoising steps~\cite{prasad2024consistency,lu2024manicm}. Offline training uses a terminal success reward and a per-step time penalty. Let $z_t=\Phi_{\mathrm{VLM}}(s_t)$ denote the observation representation and $A_t=[a_{t+i}]_{i=0}^{k-1}$ an action chunk. At each position $i$, the critic estimates $q_{t,i}=q_\phi(z_t,a_{t+i},i)$ and $v_{t,i}=v_\psi(z_t,i)$; Q is action-conditioned, whereas V depends only on the representation and chunk position.

Each sample contains an overlapping action/reward window anchored at control frame $t$. The value target is
\begin{equation}
y_{t,i}=\begin{cases}
r_{t+i}+\gamma(1-d_{t+i})v_\psi(z_t,i+1),&i<k-1,\\
r_{t+k-1}+\gamma(1-d_{t+k-1})v_\psi(z_{t+1},0),&i=k-1,
\end{cases}
\label{eq:bootstrap}
\end{equation}
where $d_{t+i}$ indicates termination and $z_{t+1}$ is the next recorded control frame, which anchors the adjacent window. We fit Q by masked squared-error regression to $y_{t,i}$ and V by IQL-style expectile regression~\cite{kostrikov2021offline} on $q_{t,i}-v_{t,i}$, using $\ell_\tau(u)=|\tau-\mathbf{1}[u<0]|u^2$ and a stop-gradient Q target. The validity mask $m_{t,i}$ includes steps through the first terminal transition, and $|M_t|=\sum_i m_{t,i}$.

Policy improvement uses a PPO-style clipped likelihood-ratio objective~\cite{schulman2017proximal} on offline demonstration actions with behavior cloning regularization. Per-position advantages are smoothed within the chunk as ~\cite{schulman2015high} $\hat A_{t,i}=(q_{t,i}-v_{t,i})+\gamma\lambda(1-d_{t+i})\hat A_{t,i+1}$, with $\hat A_{t,k}=0$. For the training likelihood, normalized actions are evaluated under a diagonal Gaussian whose mean is the consistency output and whose per-dimension standard deviation is learned. The reference likelihood is evaluated without gradients at the start of each minibatch and held fixed across its actor updates; both likelihoods use the same noisy demonstration input and noise level. Writing the current-to-reference likelihood ratio as $\rho_{t,i}$, the actor objective is
\begin{multline}
\mathcal L_{\mathrm{actor}}=\lambda_{\mathrm{bc}}\mathcal L_{\mathrm{BC}}
-\lambda_{\mathrm{ppo}}\mathbb E_{\mathcal D}\!\left[\frac{1}{|M_t|}\sum_i m_{t,i}\right.\\
\left.\cdot\min\!\left(\rho_{t,i}\hat A_{t,i},
\operatorname{clip}(\rho_{t,i},1-\epsilon,1+\epsilon)\hat A_{t,i}\right)\right].
\label{eq:ppo_full}
\end{multline}
We first warm up the actor with BC while fitting the critic, then enable the combined objective. The critic supplies advantages without propagating direct pathwise Q gradients into the actor.

\begin{figure}[t]
\centering
\includegraphics[width=0.82\columnwidth]{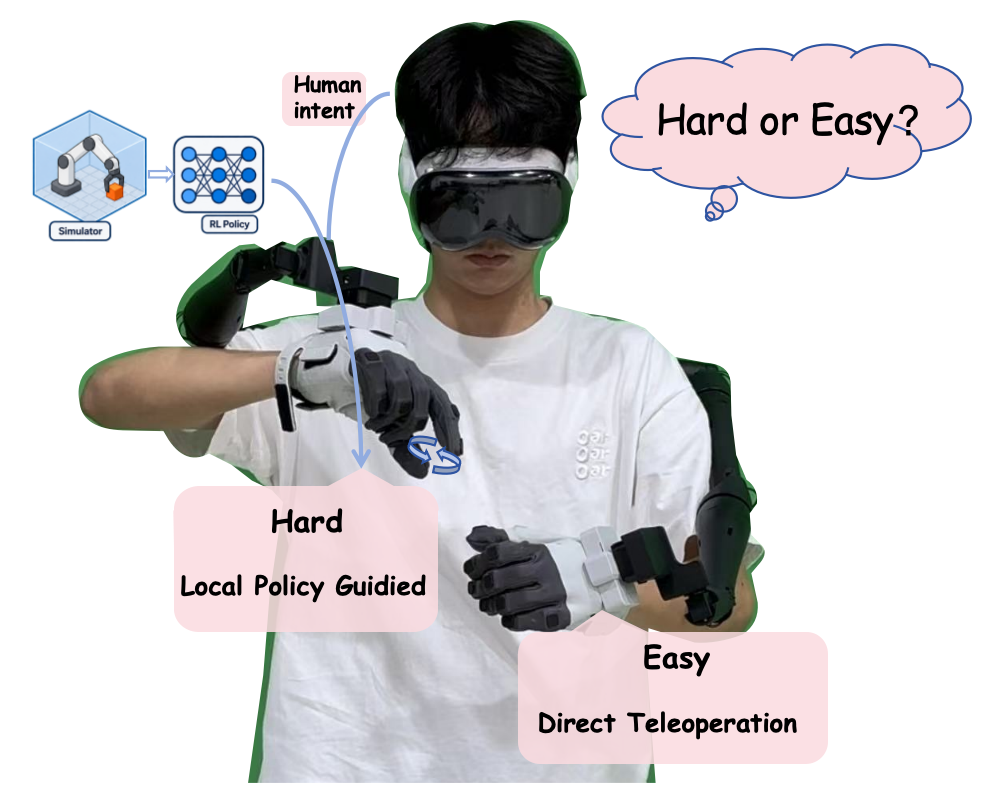}
\caption{\textbf{Two-mode human intervention system.} Wearable arm--hand motion is directly retargeted when a correction can be transferred reliably. For contact-intensive motions such as twisting, a simulation-trained local policy converts coarse human intent into coordinated, embodiment-specific finger motion. Both modes provide executable corrections for online replay.}
\label{fig:intervention_system}
\end{figure}

\paragraph{Direct and Assisted Correction}
The intervention subsystem selects between direct retargeting and local policy assistance according to the contact demands of the correction (Fig.~\ref{fig:intervention_system}). The operator monitors each rollout and initiates takeover when the policy approaches or enters a failure state. For the five single-arm tasks, we adopt the hardware and wearable teleoperation interface of DexHiL~\cite{han2026dexhil}. Human arm--hand motion is mapped to robot commands to correct failures such as incomplete finger closure or wrist misplacement, with linear interpolation smoothing the takeover transition.

\subsection{Human Intervention and Feedback}
\label{subsec:intervention}

For bimanual twisting, differences in hand geometry and actuation make direct retargeting unreliable: similar fingertip positions need not generate the forces required for rotation. We therefore add local policy assistance. The operator specifies rotation intent and timing through fingertip motion, while a robot-specific local twisting policy produces coordinated finger corrections. The operator retains the intervention decision; the local policy handles contact execution.

\paragraph{Local Policy from Human Demonstrations}
We construct the twisting reference from 15 MANUS demonstrations. Thumb and index fingertip trajectories are transformed into the object's local frame, projected onto its cylindrical surface, and aligned by normalized rotation progress. The reference comprises $K=64$ paired contact-intent nodes describing contact geometry and tangential directions. Force intent is inferred from demonstrated motion and an assumed object dynamics model rather than measured from the human hand.

A local policy for the right Linker Hand L20 is trained in simulation using asymmetric PPO. A diameter-conditioned interpolation module generates initial states across cap sizes; these states provide diverse grasp configurations and improve policy generalization across varying object geometries. The policy is optimized with rotation-progress and paired tangential-force rewards, and normalized object rotation indexes the reference node. Its actor maps the seven active joints' positions, velocities, and previous targets, together with two tangential intent directions, to seven continuous joint increments. Contact forces are available only to the simulation critic and reward; the $K$ nodes index an intent reference rather than a discrete action space. During intervention, the human provides intent and timing along the reference, and the local policy executes the corresponding correction for online data collection.

\paragraph{Feedback for Learning}
Robot interaction and human corrections are stored for online replay. For the five single-arm tasks, feedback combines terminal task success and a time penalty with a takeover penalty and an intervention-recovery bonus. Task completion and recovery from an execution error remain distinct events.

For twisting, intermediate feedback measures actual rotation during a local-policy invocation while retaining the intervention penalty. Let $\Delta\theta$ be the signed object rotation, $d\in\{-1,+1\}$ the desired direction, and $\theta^\star$ the target rotation magnitude. With $K$ reference nodes including both endpoints, the progress reward is
\begin{equation}
r_{\mathrm{prog}}=\operatorname{clip}\!\left(
\frac{d\,\Delta\theta}{\theta^\star/(K-1)},-1,1\right).
\label{eq:twisting_progress}
\end{equation}
The denominator is the average angular interval between reference nodes. Advancing by one such interval in the desired direction yields unit reward; reverse rotation yields negative feedback. This online feedback is separate from the simulation reward used to train the local policy.

\begin{figure*}[t]
\centering
\includegraphics[width=0.96\textwidth]{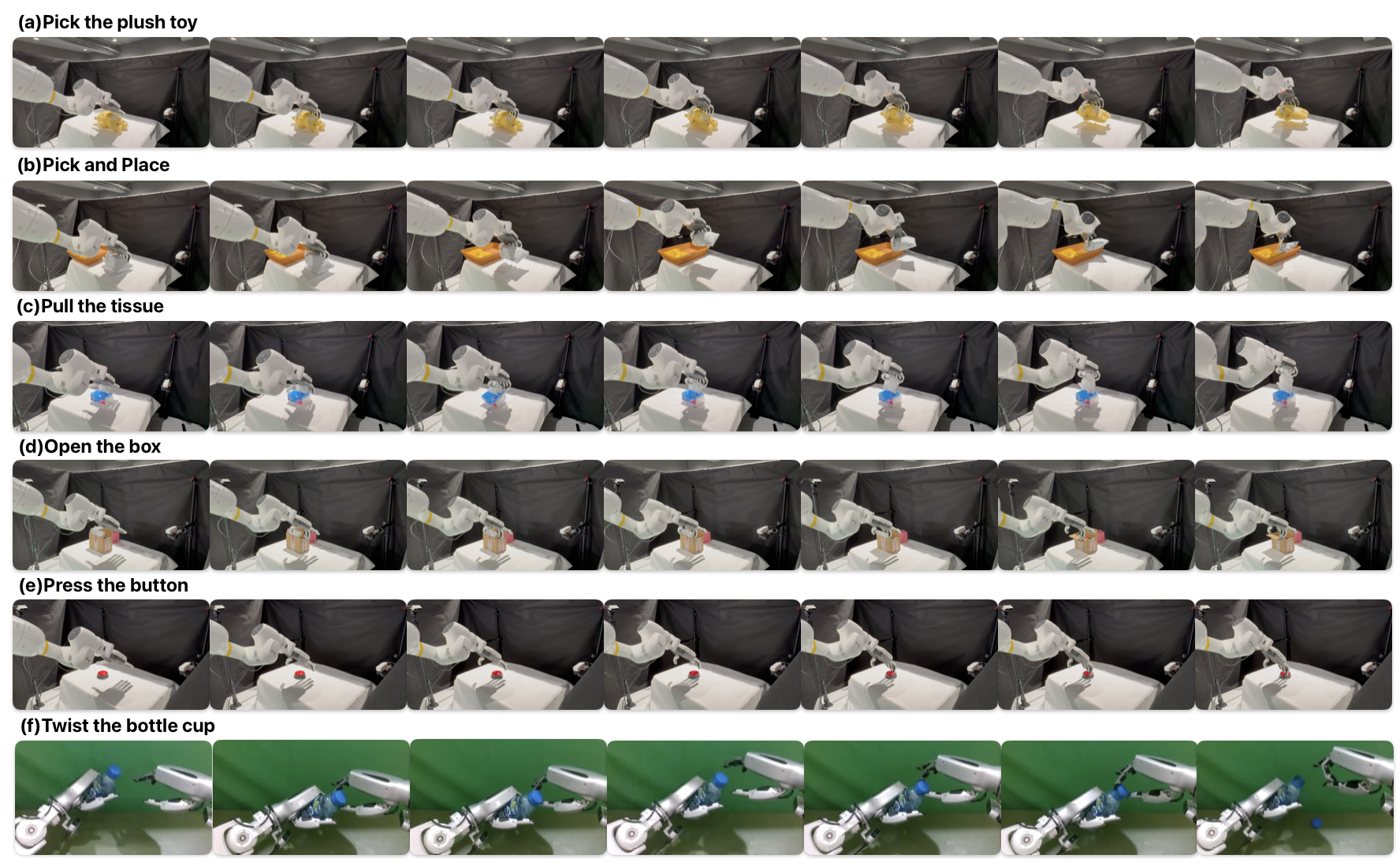}
\caption{\textbf{Execution on two robot platforms.} Rows (a)--(e) show the five tasks on the Franka--DexHand021 platform; row (f) shows bimanual cap twisting on the ROKAE--Linker Hand L20 platform. Each row presents temporally ordered frames from one rollout.}
\label{fig:experiment}
\end{figure*}

\subsection{Online Adaptation from Corrections}
\label{subsec:online_adaptation}

The online stage freezes the offline-trained VLA and learns a lightweight MLP residual actor $\pi_{\mathrm{res}}$ from robot interaction and human corrections. Given proprioception $s_{\mathrm{prop}}$, the base action chunk $A_{\mathrm{base}}$, and VLM tokens $z_t$, it produces
\begin{equation}
A_{\mathrm{final}}=A_{\mathrm{base}}+
\lambda_{\mathrm{res}}\pi_{\mathrm{res}}(s_{\mathrm{prop}},A_{\mathrm{base}},z_t),
\label{eq:residual}
\end{equation}
where $\lambda_{\mathrm{res}}$ controls correction magnitude. Across online rounds, the same residual actor continues training while the VLA base remains fixed.

The critic weights are carried into the online stage. Following RLPD~\cite{ball2023efficient}, each training batch mixes offline and online samples in a 1:1 ratio, retaining prior experience while incorporating deployment outcomes and corrections. Q and V continue training on these batches with the masked Bellman and expectile losses from Sec.~\ref{subsec:offline_rl}. To discourage residuals whose predicted value is lower than that of the base action, the actor objective includes
\begin{equation}
\mathcal L_{\mathrm{improve}}=\mathbb E\!\left[
\frac{1}{|M_t|}\sum_i m_{t,i}\max(0,q^{\mathrm{base}}_{t,i}+\delta-q^{\mathrm{final}}_{t,i})\right],
\label{eq:improve}
\end{equation}
where the two Q estimates evaluate base and composite actions at the same observation and position, and $\delta$ is a value margin. Gradients propagate through the critic to the residual actor. The complete objective is
\begin{equation}
\mathcal L_{\mathrm{res}}=\lambda_{\mathrm{imp}}\mathcal L_{\mathrm{improve}}
+\lambda_{\mathrm{bc}}\mathcal L_{\mathrm{BC}}
+\lambda_{\mathrm{mag}}\mathbb E\!\left[\|\pi_{\mathrm{res}}\|_2^2\right],
\label{eq:residual_objective}
\end{equation}
where BC is evaluated on replay actions and the final term constrains correction magnitude. The learned residual is deployed in the next round, closing the intervention--adaptation loop.

\section{Experiments}
\label{sec:experiments}

We structure the evaluation around three questions. \textbf{RQ1:} Does an integrated action-conditioned critic yield better offline policy performance than an externally attached critic? \textbf{RQ2:} Under the reported online trajectory budget, how does frozen-base residual adaptation compare with the offline policy and full-parameter, DAgger-style~\cite{ross2011reduction} updating? \textbf{RQ3:} Does local assistance increase successful execution of Twist interventions relative to direct retargeting? We evaluate all six tasks under a standard-object protocol, isolate Twist intervention execution, and test held-out objects on the five single-arm tasks.

\subsection{Platforms, Tasks, and Protocol}
\label{sec:setup}

\paragraph{Two Real-World Platforms}
The single-arm platform follows DexHiL~\cite{han2026dexhil} and comprises a Franka arm, a 12-DoF DexHand021 dexterous hand, and RealSense D435 cameras. It supports Plush Toy, Pick and Place, Open Box, Pull Tissue, and Press Button. The bimanual platform uses ROKAE arms with Linker Hand L20 dexterous hands for Twist; the local policy controls the right hand during assisted intervention. Applying the same learning pipeline to these distinct arm--hand configurations evaluates its compatibility with different hardware stacks (Fig.~\ref{fig:experiment}).

\paragraph{Tasks and Success Criteria}
The five single-arm tasks require, respectively, a stable lift off the tabletop, complete placement inside the basket, fully opening one side of the box lid, complete extraction of one tissue sheet, or full button depression without finger slippage. \textbf{Twist (Hard)} requires at least two consecutive cap-twisting motions. 

\begin{table*}[t]
\centering
\caption{\textbf{Standard-object task success.} Successful trials out of 20. DexHiL is evaluated on the three rightmost tasks; Twist uses the bimanual ROKAE--Linker Hand L20 platform, while the other tasks use Franka--DexHand021.}
\label{tab:standard}
\small
\setlength{\tabcolsep}{3pt}
\begin{tabular*}{\textwidth}{@{\extracolsep{\fill}}lcccccc@{}}
\toprule
& \multicolumn{3}{c}{Tasks without DexHiL results} & \multicolumn{3}{c}{Tasks shared with DexHiL}\\
\cmidrule(lr){2-4}\cmidrule(lr){5-7}
Method & \shortstack{Pick and\\Place} & \shortstack{Open\\Box} & \shortstack{Press\\Button} & \shortstack{Plush\\Toy} & \shortstack{Pull\\Tissue} & \shortstack{Twist\\(Hard)}\\
\midrule
VITRA & 9/20 & 12/20 & 10/20 & 14/20 & 9/20 & 6/20\\
CP Base & 10/20 & 11/20 & 13/20 & 12/20 & 7/20 & 4/20\\
Decoupled Critic (abl.) & 8/20 & 5/20 & 12/20 & 10/20 & 10/20 & 2/20\\
BORA-Offline & 14/20 & 11/20 & 13/20 & 17/20 & 12/20 & 6/20\\
BORA-Full & \textbf{18/20} & \textbf{15/20} & \textbf{17/20} & \textbf{20/20} & \textbf{16/20} & \textbf{13/20}\\
\midrule
DexHiL (full FT) & -- & -- & -- & 13/20 & 15/20 & 10/20\\
\bottomrule
\end{tabular*}

\par\vspace{10pt}
\begingroup\makeatletter\def\@captype{figure}\makeatother

\centering
\includegraphics[width=0.485\textwidth]{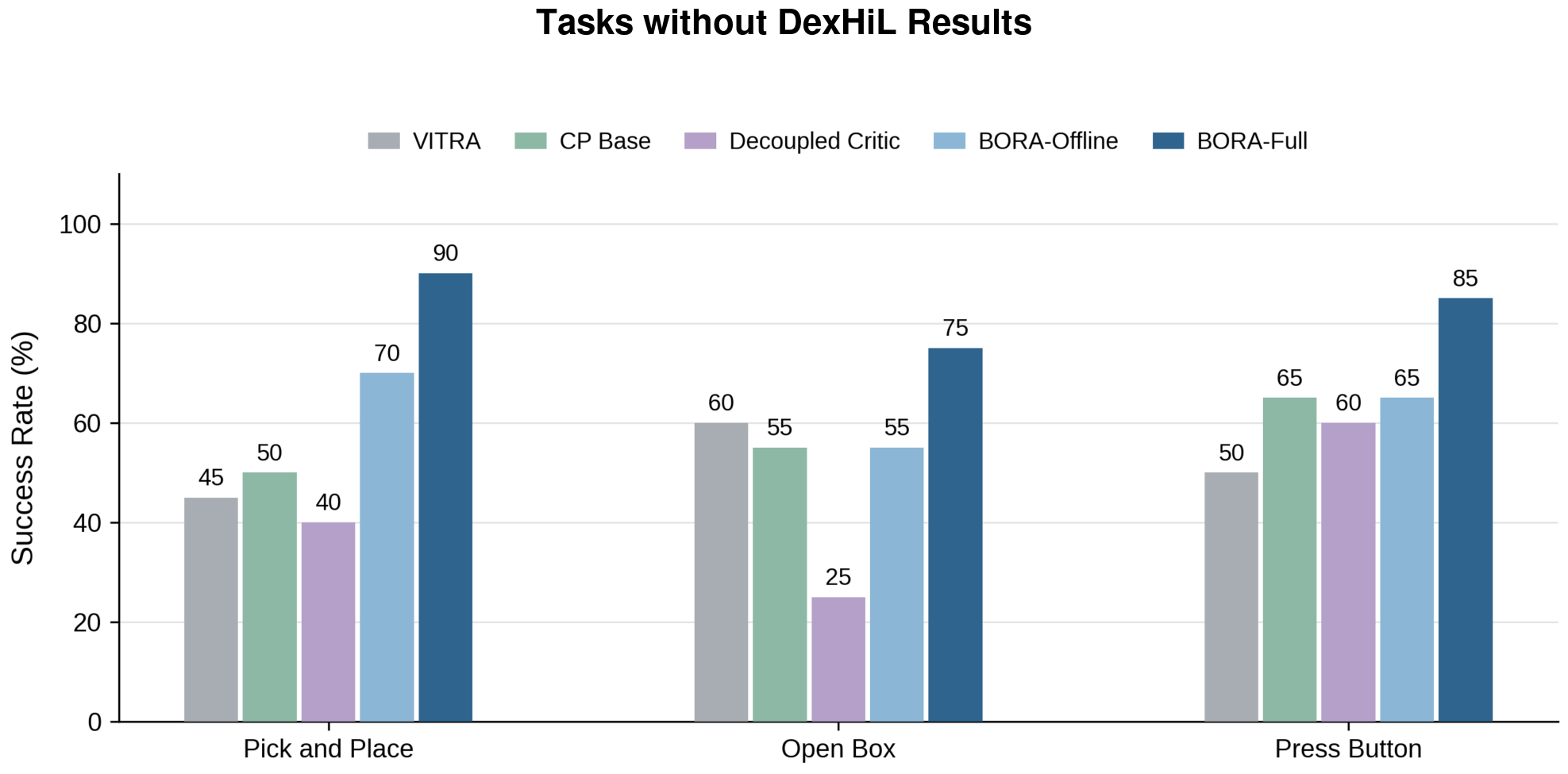}\hfill
\includegraphics[width=0.485\textwidth]{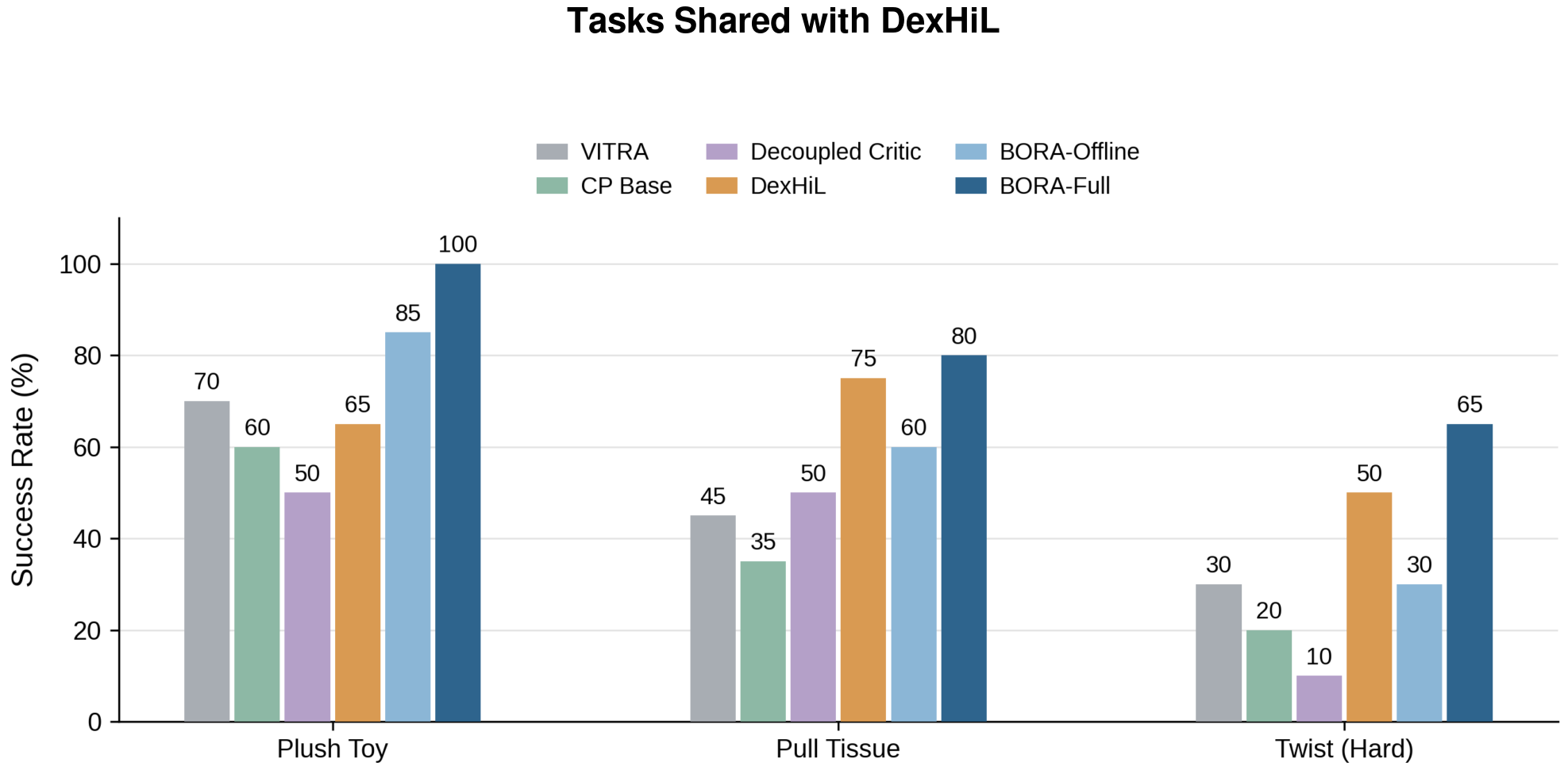}
\caption{\textbf{Standard-object success rates.} Left: tasks without DexHiL results. Right: the three tasks shared with DexHiL, including bimanual Twist (Hard). Each bar reports the observed success rate over 20 trials; exact counts are given in Table~\ref{tab:standard}.}
\label{fig:standard_success}

\endgroup
\end{table*}

\paragraph{Data and Evaluation}
The offline datasets contain 60 trajectories for Plush Toy, 100 each for Pick and Place and Open Box, 60 each for Pull Tissue and Press Button, and 100 for the bimanual Twist task. All methods use the same task-specific offline data, pretrained initialization, and evaluation protocol wherever their task coverage overlaps. Each online method performs two adaptation rounds with 10 trajectories per task and round, adding 20 online trajectories per task. We match the trajectory budget rather than the intervention-segment count: takeover occurs only when a policy reaches its own failure state, so intervention timing and frequency are method-dependent. On the three tasks shared with DexHiL, both methods use the same offline data, online trajectory budget, and test conditions.

Online rollouts use training objects, and evaluation is conducted separately from data collection and updates. Each method is tested for 20 trials per reported task and setting. Standard tests use object instances represented in training; held-out-object tests use instances excluded from both offline training and online adaptation. The latter setting covers the five Franka--DexHand021 tasks.

\paragraph{Baselines and Variants}
All variants start from the same pretrained VITRA VLM backbone. \textbf{VITRA}~\cite{li2025scalable} is an imitation-learning baseline with a diffusion action expert, and \textbf{CP Base} replaces it with a consistency policy without RL post-training. \textbf{Decoupled Critic}~\cite{luo2025precise} is an architectural ablation that pairs the VLA with a separately trained image-and-state critic that does not share the actor's cognition tokens. \textbf{BORA-Offline} instead evaluates the actor's cognition tokens and action chunks jointly. \textbf{BORA-Full} adds two rounds of frozen-base, human-guided residual adaptation. \textbf{DexHiL}~\cite{han2026dexhil} serves as the full-parameter online adaptation baseline on Plush Toy, Pull Tissue, and Twist.

\subsection{System Performance and Human-Guided Adaptation}
\label{sec:main_exp}

\paragraph{Overall Performance}
Across the six standard-object tasks (Table~\ref{tab:standard} and Fig.~\ref{fig:standard_success}), BORA-Full records 99/120 successes (82.5\%), compared with 73/120 (60.8\%) for BORA-Offline, 60/120 (50.0\%) for VITRA, 57/120 (47.5\%) for CP Base, and 47/120 (39.2\%) for the decoupled-critic ablation. BORA-Full has the highest observed count on every task across both platforms. Local assistance is used only for Twist; the five Franka--DexHand021 tasks use direct wearable teleoperation.

\paragraph{Integrated versus Decoupled Value Guidance (RQ1)}
Relative to the externally attached critic, BORA-Offline records higher observed success on all six standard tasks (73/120 versus 47/120) and a higher held-out-object mean (52\% versus 41\%); the largest gap is on Open Box (7/20 versus 1/20). The decoupled critic nevertheless matches BORA-Offline on two held-out tasks, so its guidance is less consistent rather than uniformly ineffective. CP Base records 57/120 successes compared with 60/120 for VITRA, showing that consistency generation alone does not account for BORA-Offline's 73/120 result. These downstream comparisons are consistent with a benefit from jointly evaluating VLM cognition tokens and generated action chunks, but they do not constitute a critic-calibration test.

\paragraph{Combined Effect of Online Data and Residual Adaptation (RQ2)}
With 20 additional trajectories per task, BORA-Full records 99/120 successes versus 73/120 for BORA-Offline, an observed increase of 21.7 percentage points. This contrast measures the combined effect of online experience and residual updating rather than a residual-only effect. Gains occur on both hardware stacks and range from three to seven additional successes per task; the largest is on Twist, from 6/20 to 13/20.

\begin{table*}[t]
\centering
\caption{\textbf{Generalization to held-out objects on five single-arm tasks.} Successful trials out of 20; BORA-Full's online data use only training objects, and the reported objects are excluded from both learning stages.}
\label{tab:unseen}
\small
\setlength{\tabcolsep}{8pt}
\begin{tabular}{@{}lcccccc@{}}
\toprule
Method & Plush Toy & Pick and Place & Open Box & Pull Tissue & Press Button & Mean (\%)\\
\midrule
VITRA & 9/20 & 6/20 & 2/20 & 7/20 & 9/20 & 33\\
CP Base & 7/20 & 6/20 & 3/20 & 3/20 & 8/20 & 27\\
Decoupled Critic (abl.) & 11/20 & 8/20 & 1/20 & 10/20 & 11/20 & 41\\
BORA-Offline & 15/20 & 9/20 & 7/20 & 10/20 & 11/20 & 52\\
BORA-Full & \textbf{17/20} & \textbf{14/20} & \textbf{10/20} & \textbf{14/20} & \textbf{15/20} & \textbf{70}\\
\bottomrule
\end{tabular}
\end{table*}

\paragraph{Comparison with Full-Parameter DexHiL (RQ2)}
On the three tasks shared with DexHiL, BORA-Full records 49/60 observed successes (81.7\%) versus 38/60 (63.3\%) under the matched protocol in Sec.~\ref{sec:setup}. The task-level counts are 20/20 versus 13/20 on Plush Toy, 16/20 versus 15/20 on Pull Tissue, and 13/20 versus 10/20 on Twist, indicating that the observed improvement is not uniform across tasks.

DexHiL aggregates intervention trajectories and fully fine-tunes the VLA using an intervention-aware imitation objective, whereas BORA preserves the offline policy and learns a compact residual under value guidance. Twist illustrates an important distinction between reproducing corrective actions and optimizing their outcomes. After losing effective contact with the cap, an imitation-based policy may still replay the nominal two-step twisting motion without producing the required cap rotation, because action agreement alone does not explicitly penalize the absence of object progress. In contrast, BORA trains its action-conditioned critic with outcome-based rewards that include rotation progress. Action chunks associated with slippage or ineffective rotation can therefore receive lower values, allowing the residual policy to favor corrections associated with maintaining or recovering effective contact. This interpretation is consistent with BORA's stronger observed Twist performance, but the comparison remains system-level rather than a learning-rule-only ablation: intervention segments are policy-dependent, and BORA additionally uses policy-assisted teleoperation. Table~\ref{tab:twist_intervention} separately evaluates this intervention executability. Under our implementations and training settings, DexHiL uses four H100 GPUs for full-model adaptation, whereas one BORA update completes in approximately one hour on a single RTX~4090; this reports implementation cost rather than a hardware-normalized speedup.

\begin{table}[t]
\centering
\caption{\textbf{Local-policy assistance for Twist intervention.} Successful execution of the requested correction over 20 intervention attempts per condition.}
\label{tab:twist_intervention}
\small
\setlength{\tabcolsep}{5pt}
\begin{tabular}{@{}lcc@{}}
\toprule
Intervention mode & Successes & Success rate\\
\midrule
Direct retargeting (no assistance) & 6/20 & 30\%\\
Local-policy-assisted (ours) & \textbf{20/20} & \textbf{100\%}\\
\bottomrule
\end{tabular}
\end{table}

\paragraph{Executable Intervention for Twisting (RQ3)}
Table~\ref{tab:twist_intervention} isolates correction execution from subsequent policy learning. Direct retargeting succeeds in 6/20 interventions (30\%), whereas local assistance succeeds in 20/20. The failed direct takeovers are consistent with the human--robot embodiment gap: the L20 thumb differs from the human thumb in base position, reachable contact geometry, and motion pattern. Pose retargeting can therefore place fingertips near the desired locations without establishing the opposing contacts or tangential motion required for rotation. By mapping operator rotation intent to robot-specific finger coordination, the local policy removes the need for direct finger-by-finger control and produces executable corrections for online adaptation.

\paragraph{System-Level Interpretation}
The three comparisons clarify the role of each component. The decoupled-critic ablation tests whether value guidance can simply be attached to a VLA; its lower observed success motivates sharing the policy representation and action chunks. BORA-Full then tests whether deployment experience and human corrections can improve this offline policy without modifying the VLA base. Finally, the Twist study tests whether assistance executes corrections more successfully than direct retargeting under the evaluated conditions. The components therefore address different failure modes: the critic ranks candidate behavior, the residual actor limits the scope of online change, and the intervention subsystem determines the quality of corrective experience entering replay.

\begin{figure}[b]
\centering
\includegraphics[width=\columnwidth]{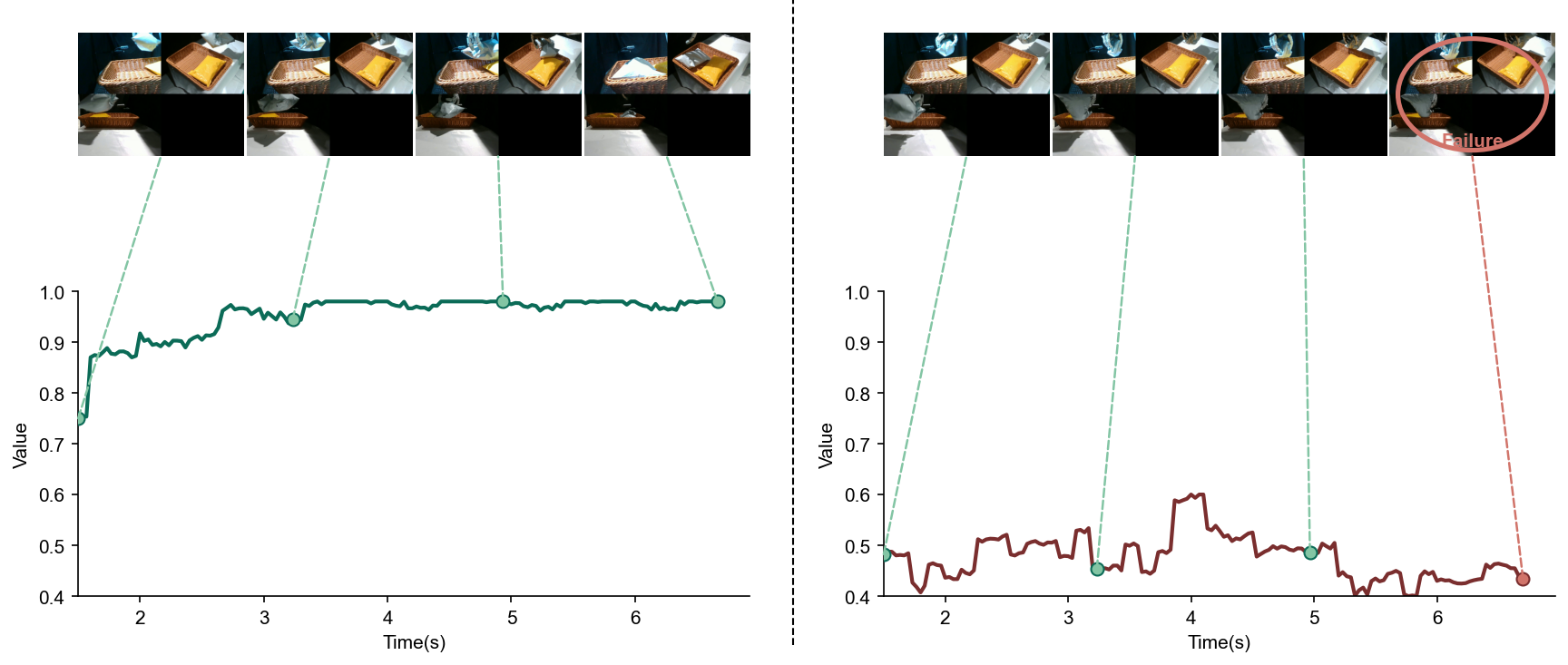}
\caption{\textbf{Representative critic value profiles.} In this example, the successful rollout (left) has higher predicted values than the failed rollout (right); dashed lines associate observations with the plotted estimates.}
\label{fig:value_vis}
\end{figure}
\subsection{Generalization to Held-Out Objects}

On objects excluded from both learning stages, average success is 27\% for CP Base, 33\% for VITRA, 41\% for the decoupled critic, 52\% for BORA-Offline, and 70\% for BORA-Full (Table~\ref{tab:unseen}). CP Base drops most from standard-object performance, while BORA-Offline remains 25 percentage points higher. This ordering is consistent with VLA-grounded value guidance retaining task structure under object variation.

\begin{figure}[tb]
\centering
\includegraphics[width=\columnwidth]{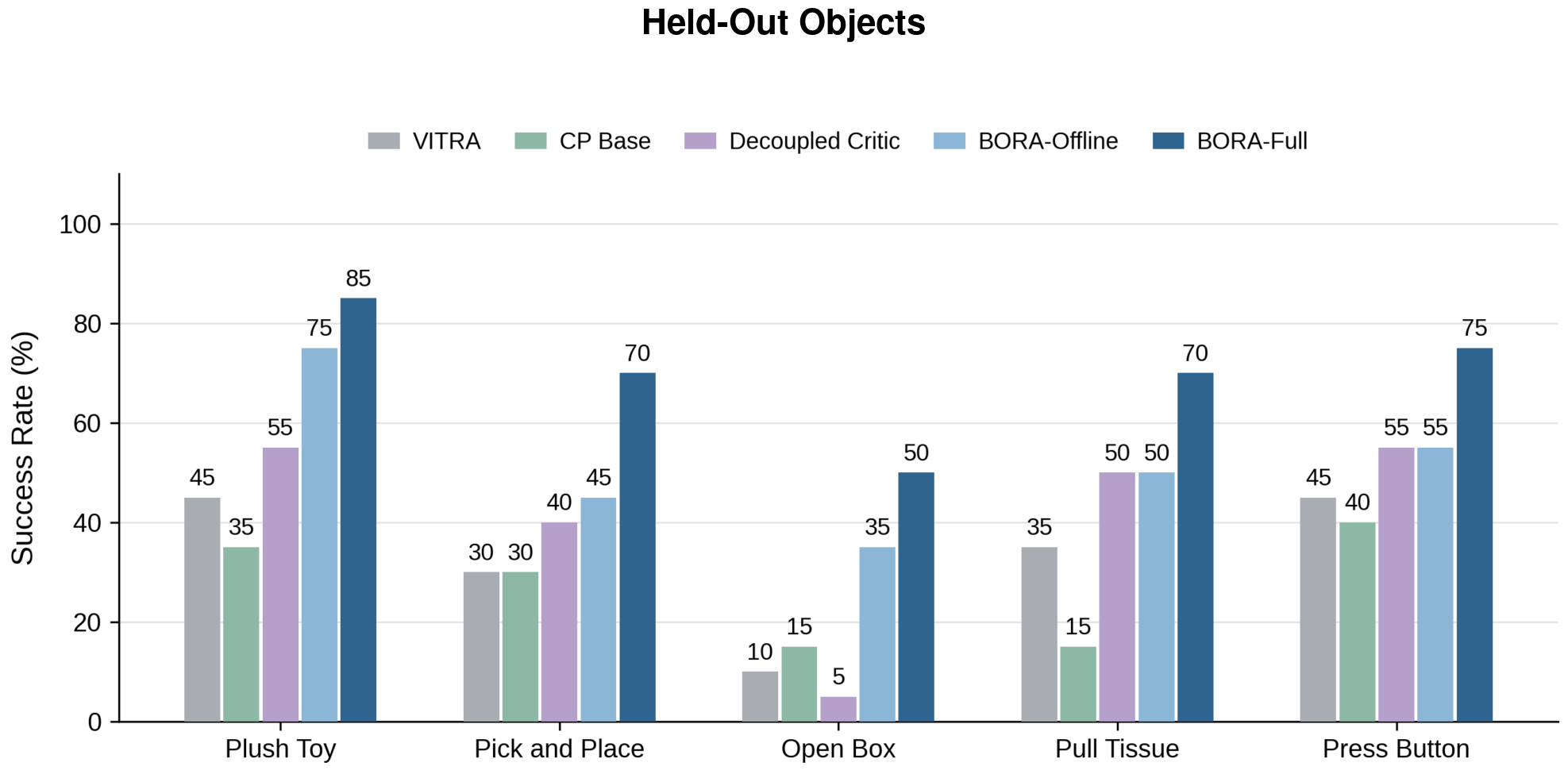}
\caption{\textbf{Held-out-object success rates.} Percentages on the five Franka--DexHand021 tasks using objects excluded from both learning stages; see Table~\ref{tab:unseen}.}
\label{fig:unseen_success}
\end{figure}

Relative to BORA-Offline, BORA-Full gains 18 percentage points on average: +10 on Plush Toy, +25 on Pick and Place, +15 on Open Box, and +20 on both Pull Tissue and Press Button (Fig.~\ref{fig:unseen_success}). All corrections are collected with training objects, so these gains reflect transfer to unseen instances rather than direct exposure to the evaluation objects. Their consistency across tasks accords with the residual actor correcting recurring execution errors while the frozen VLA retains its visual prior.

\paragraph{Qualitative Value Estimates}
Figure~\ref{fig:value_vis} contrasts a representative successful rollout, whose value rises and remains high, with a failed rollout whose value falls near failure. Because corrections concentrate near visually similar pre-failure states, the critic must distinguish whether an action chunk preserves contact rather than merely detecting the target object. Together with the decoupled-critic ablation, these profiles suggest that inherited action-conditioned values remain informative online; they are illustrative, not a calibration or causal test of critic inheritance.

\section{Limitations}
\label{sec:limitations}
BORA relies on visual and proprioceptive inputs without dense tactile feedback. Although we instantiate the pipeline on two platforms and hand models, different tasks are assigned to each; the results demonstrate system compatibility, not transfer of one policy across embodiments.Finally, Held-out-object evaluation covers only the five single-arm tasks, and local assistance is tested only for thumb--index twisting using a demonstration-derived reference.

\section{Conclusion}
\label{sec:conclusion}
BORA presents an offline-to-online reinforcement learning framework for real-world dexterous VLA adaptation. It combines policy-aligned value guidance, executable high-DoF human correction, and lightweight residual learning while preserving the pretrained base policy. Experiments across six tasks and two arm--hand platforms demonstrate improved performance and generalization under a limited online interaction budget, together with more reliable intervention for bimanual twisting. These results provide a practical path toward robust and efficient real-world dexterous manipulation.

\bibliographystyle{IEEEtran}
\bibliography{references}

@article{kim2024openvla,
  title={Openvla: An open-source vision-language-action model},
  author={Kim, Moo Jin and Pertsch, Karl and Karamcheti, Siddharth and Xiao, Ted and Balakrishna, Ashwin and Nair, Suraj and Rafailov, Rafael and Foster, Ethan and Lam, Grace and Sanketi, Pannag and others},
  journal={arXiv preprint arXiv:2406.09246},
  year={2024}
}

@article{black2024pi_0,
  title={{$\pi_0$}: A Vision-Language-Action Flow Model for General Robot Control},
  author={Black, Kevin and Brown, Noah and Driess, Danny and Esmail, Adnan and Equi, Michael and Finn, Chelsea and Fusai, Niccolo and Groom, Lachy and Hausman, Karol and Ichter, Brian and others},
  journal={arXiv preprint arXiv:2410.24164},
  year={2024}
}

@article{intelligence2025pi_,
  title={{$\pi_{0.5}$}: A Vision-Language-Action Model with Open-World Generalization},
  author={{Physical Intelligence} and Black, Kevin and Brown, Noah and Darpinian, James and Dhabalia, Karan and Driess, Danny and Esmail, Adnan and Equi, Michael and Finn, Chelsea and Fusai, Niccolo and others},
  journal={arXiv preprint arXiv:2504.16054},
  year={2025}
}

@article{luo2026being,
  title={Being-H0. 5: Scaling Human-Centric Robot Learning for Cross-Embodiment Generalization},
  author={Luo, Hao and Wang, Ye and Zhang, Wanpeng and Zheng, Sipeng and Xi, Ziheng and Xu, Chaoyi and Xu, Haiweng and Yuan, Haoqi and Zhang, Chi and Wang, Yiqing and others},
  journal={arXiv preprint arXiv:2601.12993},
  year={2026}
}

@article{zhong2025dexgraspvla,
  title={Dexgraspvla: A vision-language-action framework towards general dexterous grasping},
  author={Zhong, Yifan and Huang, Xuchuan and Li, Ruochong and Zhang, Ceyao and Chen, Zhang and Guan, Tianrui and Zeng, Fanlian and Lui, Ka Num and Ye, Yuyao and Liang, Yitao and others},
  journal={arXiv preprint arXiv:2502.20900},
  year={2025}
}

@inproceedings{he2025dexvlg,
  title={Dexvlg: Dexterous vision-language-grasp model at scale},
  author={He, Jiawei and Li, Danshi and Yu, Xinqiang and Qi, Zekun and Zhang, Wenyao and Chen, Jiayi and Zhang, Zhaoxiang and Zhang, Zhizheng and Yi, Li and Wang, He},
  booktitle={Proceedings of the IEEE/CVF International Conference on Computer Vision},
  pages={14248--14258},
  year={2025}
}

@article{li2025scalable,
  title={Scalable vision-language-action model pretraining for robotic manipulation with real-life human activity videos},
  author={Li, Qixiu and Deng, Yu and Liang, Yaobo and Luo, Lin and Zhou, Lei and Yao, Chengtang and Zeng, Lingqi and Feng, Zhiyuan and Liang, Huizhi and Xu, Sicheng and others},
  journal={arXiv preprint arXiv:2510.21571},
  year={2025}
}

@article{luo2025being,
  title={Being-h0: vision-language-action pretraining from large-scale human videos},
  author={Luo, Hao and Feng, Yicheng and Zhang, Wanpeng and Zheng, Sipeng and Wang, Ye and Yuan, Haoqi and Liu, Jiazheng and Xu, Chaoyi and Jin, Qin and Lu, Zongqing},
  journal={arXiv preprint arXiv:2507.15597},
  year={2025}
}

@article{chen2025conrft,
  title={Conrft: A reinforced fine-tuning method for vla models via consistency policy},
  author={Chen, Yuhui and Tian, Shuai and Liu, Shugao and Zhou, Yingting and Li, Haoran and Zhao, Dongbin},
  journal={arXiv preprint arXiv:2502.05450},
  year={2025}
}

@article{huang2025co,
  title={Co-rft: Efficient fine-tuning of vision-language-action models through chunked offline reinforcement learning},
  author={Huang, Dongchi and Fang, Zhirui and Zhang, Tianle and Li, Yihang and Zhao, Lin and Xia, Chunhe},
  journal={arXiv preprint arXiv:2508.02219},
  year={2025}
}

@article{xu2026rl,
  title={RL Token: Bootstrapping Online RL with Vision-Language-Action Models},
  author={Xu, Charles and Springenberg, Jost Tobias and Equi, Michael and Amin, Ali and Esmail, Adnan and Levine, Sergey and Ke, Liyiming},
  journal={arXiv preprint arXiv:2604.23073},
  year={2026}
}

@article{chen2025pirl,
  title={$\pi$RL: Online rl fine-tuning for flow-based vision-language-action models},
  author={Chen, Kang and Liu, Zhihao and Zhang, Tonghe and Guo, Zhen and Xu, Si and Lin, Hao and Zang, Hongzhi and Zhang, Quanlu and Yu, Zhaofei and Fan, Guoliang and others},
  journal={arXiv preprint arXiv:2510.25889},
  year={2025}
}

@article{nakamoto2023cal,
  title={Cal-ql: Calibrated offline rl pre-training for efficient online fine-tuning},
  author={Nakamoto, Mitsuhiko and Zhai, Simon and Singh, Anikait and Sobol Mark, Max and Ma, Yi and Finn, Chelsea and Kumar, Aviral and Levine, Sergey},
  journal={Advances in Neural Information Processing Systems},
  volume={36},
  pages={62244--62269},
  year={2023}
}

@inproceedings{ball2023efficient,
  title={Efficient online reinforcement learning with offline data},
  author={Ball, Philip J and Smith, Laura and Kostrikov, Ilya and Levine, Sergey},
  booktitle={International Conference on Machine Learning},
  pages={1577--1594},
  year={2023},
  organization={PMLR}
}

@article{lu2024manicm,
  title={Manicm: Real-time 3d diffusion policy via consistency model for robotic manipulation},
  author={Lu, Guanxing and Gao, Zifeng and Chen, Tianxing and Dai, Wenxun and Wang, Ziwei and Ding, Wenbo and Tang, Yansong},
  journal={arXiv preprint arXiv:2406.01586},
  year={2024}
}

@inproceedings{ankile2025imitation,
  title={From imitation to refinement-residual rl for precise assembly},
  author={Ankile, Lars and Simeonov, Anthony and Shenfeld, Idan and Torne, Marcel and Agrawal, Pulkit},
  booktitle={2025 IEEE International Conference on Robotics and Automation (ICRA)},
  pages={01--08},
  year={2025},
  organization={IEEE}
}

@inproceedings{zhou2025efficient,
  title={Efficient online reinforcement learning fine-tuning need not retain offline data},
  author={Zhou, Zhiyuan and Peng, Andy and Li, Qiyang and Levine, Sergey and Kumar, Aviral},
  booktitle={International Conference on Learning Representations},
  volume={2025},
  pages={32343--32368},
  year={2025}
}

@article{han2026dexhil,
  title={DexHiL: A Human-in-the-Loop Framework for Vision-Language-Action Model Post-Training in Dexterous Manipulation},
  author={Han, Yifan and Chen, Zhongxi and Zhao, Yuxuan and Xu, Congsheng and Shao, Yanming and Peng, Yichuan and Mu, Yao and Lian, Wenzhao},
  journal={arXiv preprint arXiv:2603.09121},
  year={2026}
}

@inproceedings{su2026rfs,
  title={Rfs: Reinforcement learning with residual flow steering for dexterous manipulation},
  author={Su, Entong and Westenbroek, Tyler and Nagabandi, Anusha and Gupta, Abhishek},
  booktitle={The Fourteenth International Conference on Learning Representations},
  year={2026}
}

@article{prasad2024consistency,
  title={Consistency policy: Accelerated visuomotor policies via consistency distillation},
  author={Prasad, Aaditya and Lin, Kevin and Wu, Jimmy and Zhou, Linqi and Bohg, Jeannette},
  journal={arXiv preprint arXiv:2405.07503},
  year={2024}
}

@article{mandlekar2021matters,
  title={What matters in learning from offline human demonstrations for robot manipulation},
  author={Mandlekar, Ajay and Xu, Danfei and Wong, Josiah and Nasiriany, Soroush and Wang, Chen and Kulkarni, Rohun and Fei-Fei, Li and Savarese, Silvio and Zhu, Yuke and Mart{\'\i}n-Mart{\'\i}n, Roberto},
  journal={arXiv preprint arXiv:2108.03298},
  year={2021}
}

@inproceedings{zitkovich2023rt,
  title={Rt-2: Vision-language-action models transfer web knowledge to robotic control},
  author={Zitkovich, Brianna and Yu, Tianhe and Xu, Sichun and Xu, Peng and Xiao, Ted and Xia, Fei and Wu, Jialin and Wohlhart, Paul and Welker, Stefan and Wahid, Ayzaan and others},
  booktitle={Conference on Robot Learning},
  pages={2165--2183},
  year={2023},
  organization={PMLR}
}

@inproceedings{mees2024octo,
  title={Octo: An open-source generalist robot policy},
  author={Mees, Oier and Ghosh, Dibya and Pertsch, Karl and Black, Kevin and Walke, Homer Rich and Dasari, Sudeep and Hejna, Joey and Kreiman, Tobias and Xu, Charles and Luo, Jianlan and others},
  booktitle={First Workshop on Vision-Language Models for Navigation and Manipulation at ICRA 2024},
  year={2024}
}

@article{chen2025flowing,
  title={Flowing from reasoning to motion: Learning 3d hand trajectory prediction from egocentric human interaction videos},
  author={Chen, Mingfei and Wang, Yifan and Li, Zhengqin and Bharadhwaj, Homanga and Chen, Yujin and Qin, Chuan and Kou, Ziyi and Tian, Yuan and Whitmire, Eric and Sodhi, Rajinder and others},
  journal={arXiv preprint arXiv:2512.16907},
  year={2025}
}

@inproceedings{yin2025geometric,
  title={Geometric retargeting: A principled, ultrafast neural hand retargeting algorithm},
  author={Yin, Zhao-Heng and Wang, Changhao and Pineda, Luis and Bodduluri, Krishna and Wu, Tingfan and Abbeel, Pieter and Mukadam, Mustafa},
  booktitle={2025 IEEE/RSJ International Conference on Intelligent Robots and Systems (IROS)},
  pages={17376--17382},
  year={2025},
  organization={IEEE}
}

@article{zhu2026learning,
  title={Learning dexterous manipulation using contact wrench guidance from human demonstration},
  author={Zhu, Xinghao and Liu, Zixi and Jain, Shalin and Li, Chenran and Noori, Milad and Lin, Michael Andres and Zhao, Huihua and Welsh, John and Verghese, Mrinal and Liu, Wei and others},
  journal={arXiv preprint arXiv:2607.00033},
  year={2026}
}

@article{li2026towards,
  title={Towards Human-level Dexterous Teleoperation},
  author={Li, Puhao and Chen, Zeyuan and Wu, Yingying and Wei, Pengkun and Li, Yuyang and Wang, Tianyu and Shi, Jiaxiao and Yu, Mingrui and Jia, Baoxiong and Zhu, Song-chun and others},
  journal={arXiv preprint arXiv:2607.11481},
  year={2026}
}

@article{yin2025dexteritygen,
  title={Dexteritygen: Foundation controller for unprecedented dexterity},
  author={Yin, Zhao-Heng and Wang, Changhao and Pineda, Luis and Hogan, Francois and Bodduluri, Krishna and Sharma, Akash and Lancaster, Patrick and Prasad, Ishita and Kalakrishnan, Mrinal and Malik, Jitendra and others},
  journal={arXiv preprint arXiv:2502.04307},
  year={2025}
}

@article{lei2025rl,
  title={Rl-100: Performant robotic manipulation with real-world reinforcement learning},
  author={Lei, Kun and Li, Huanyu and Yu, Dongjie and Wei, Zhenyu and Guo, Lingxiao and Jiang, Zhennan and Wang, Ziyu and Liang, Shiyu and Xu, Huazhe},
  journal={arXiv preprint arXiv:2510.14830},
  year={2025}
}

@article{kostrikov2021offline,
  title={Offline reinforcement learning with implicit q-learning},
  author={Kostrikov, Ilya and Nair, Ashvin and Levine, Sergey},
  journal={arXiv preprint arXiv:2110.06169},
  year={2021}
}

@article{schulman2017proximal,
  title={Proximal policy optimization algorithms},
  author={Schulman, John and Wolski, Filip and Dhariwal, Prafulla and Radford, Alec and Klimov, Oleg},
  journal={arXiv preprint arXiv:1707.06347},
  year={2017}
}

@article{schulman2015high,
  title={High-dimensional continuous control using generalized advantage estimation},
  author={Schulman, John and Moritz, Philipp and Levine, Sergey and Jordan, Michael and Abbeel, Pieter},
  journal={arXiv preprint arXiv:1506.02438},
  year={2015}
}

@article{luo2025precise,
  title={Precise and dexterous robotic manipulation via human-in-the-loop reinforcement learning},
  author={Luo, Jianlan and Xu, Charles and Wu, Jeffrey and Levine, Sergey},
  journal={Science Robotics},
  volume={10},
  number={105},
  pages={eads5033},
  year={2025},
  publisher={American Association for the Advancement of Science}
}

@inproceedings{ross2011reduction,
  title={A reduction of imitation learning and structured prediction to no-regret online learning},
  author={Ross, St{\'e}phane and Gordon, Geoffrey and Bagnell, Drew},
  booktitle={Proceedings of the fourteenth international conference on artificial intelligence and statistics},
  pages={627--635},
  year={2011},
  organization={JMLR Workshop and Conference Proceedings}
}

@article{mandlekar2020human,
  title={Human-in-the-loop imitation learning using remote teleoperation},
  author={Mandlekar, Ajay and Xu, Danfei and Mart{\'\i}n-Mart{\'\i}n, Roberto and Zhu, Yuke and Fei-Fei, Li and Savarese, Silvio},
  journal={arXiv preprint arXiv:2012.06733},
  year={2020}
}
\end{document}